\documentclass{article}

\usepackage{arxiv}

\usepackage[utf8]{inputenc} % allow utf-8 input
\usepackage[T1]{fontenc}    % use 8-bit T1 fonts
\usepackage{hyperref}       % hyperlinks
\usepackage{url}            % simple URL typesetting
\usepackage{booktabs}       % professional-quality tables
\usepackage{amsfonts}       % blackboard math symbols
\usepackage{nicefrac}       % compact symbols for 1/2, etc.
\usepackage{microtype}      % microtypography
\usepackage{lipsum}
\usepackage{graphics} % for pdf, bitmapped graphics files
\usepackage{epsfig} % for postscript graphics files
\usepackage{mathptmx} % assumes new font selection scheme installed
\usepackage{amsmath} % assumes amsmath package installed
\usepackage{amssymb}  % assumes amsmath package installed

\title{Unsupervised anomaly detection for a Smart Autonomous Robotic
Assistant Surgeon (SARAS) using a deep residual autoencoder}

\author{
  Dinesh Jackson Samuel \\
  Postdoctoral Researcher\\
  Faculty of Technology, Design and Environment\\
  Visual Artificial Intelligence Laboratory\\
  Oxford Brookes University \\
  \texttt{rsamuel@brookes.ac.uk, jacksoncse@gmail.com} \\
   \And
 Fabio Cuzzolin \\
 Professor\\
  Faculty of Technology, Design and Environment\\
  Head of Visual Artificial Intelligence Laboratory\\
  Oxford Brookes University \\
  \texttt{fabio.cuzzolin@brookes.ac.uk} \\
}

\begin{document}
\maketitle

\begin{abstract}
%\lipsum[1]
Anomaly detection in Minimally-Invasive Surgery
(MIS) traditionally requires a human expert monitoring the
procedure from a console. Data scarcity, on the other hand, hinders what would be a desirable migration towards autonomous robotic-assisted surgical systems. Automated anomaly detection systems in this area typically rely on classical supervised
learning. Anomalous events in a surgical setting, however, are rare, making it difficult to capture data to train a detection model in a supervised fashion. In this work we thus propose an unsupervised approach to anomaly detection for robotic-assisted surgery based on deep residual autoencoders. The idea is to make the autoencoder learn the ’normal’ distribution of the data and detect abnormal events deviating from this distribution by measuring the reconstruction error. The model is trained and validated upon both the publicly available Cholec80
dataset, provided with extra annotation, and on a set of videos captured on procedures using artificial anatomies (’phantoms’) produced as part of the Smart Autonomous Robotic Assistant Surgeon (SARAS) project. The system achieves recall and
precision equal to 78.4\%, 91.5\%, respectively, on Cholec80 and of 95.6\%, 88.1\% on the SARAS phantom dataset. The end-to-end system was developed and deployed as part of the SARAS demonstration platform for real-time anomaly detection with a
processing time of about 25 ms per frame.
\end{abstract}

% keywords can be removed

\section{INTRODUCTION}
In recent years, Minimally-Invasive Surgery (MIS) has been strongly gaining ground. Endoscopic surgery in general has attracted a great deal of interest, as it only requires small incisions (5-30 mm) to give the endoscope and other instruments access to the surgical cavity, rather than the vast incision (approximately 300 mm) demanded by traditional surgery. Endoscopic surgery does therefore result in shorter
recovery times compared to ’open’ surgery. The future of robotic surgery is arguably driven by a combination of advances in computer vision and MIS, with the goal of
minimising human intervention. In recent years, robotic MIS (R-MIS) has been proven effective in the surgical theatre under the supervision of expert surgeons. Although
this technology is adaptive, precise and accurate, most R-MIS systems are not designed to replace the main surgeon conducting the procedure but to increase the safety and effectiveness of surgeries. One such kind of human-machine interactive robotic system, named “da Vinci”, has been developed by Intuitive Surgical to perform precise and complex surgeries through small incisions
\cite{Guthart2000TheIT}. Medical associations have later approved this system for assisting otolaryngologists with ENT and prostatectomy surgical procedures, including endoscopy \cite{Vitiello2013EmergingRP}. Generally, robotic-assisted surgical techniques have the potential to
overcome human errors, delivering high precision, reliability, and accuracy under human supervision. Surgical procedures can be very long, exhausting and cumbersome, leading to fatigue and hand trembling \cite{LEE201033}. A surgeon’s extended workload, involving manipulation, visualisation and mental stress, may significantly affect their performance.

The typical surgical environment encompasses a patient table, a main surgeon, two assistant surgeons and two nurses. The assistant surgeon plays a key role both before and during the surgery. When using da Vinci, the main surgeon monitors
and controls the robotic arm from the endoscopic console, while the assistant surgeon abets the da Vinci in handling the tools. These robots are not autonomous and can be considered as mere extensions of the main surgeon. Crucially, the assistant surgeon is active for only 30\% of the time and remains idle during the rest of the surgery. As a result, using a da Vinci does not alleviate human surgeons’ schedules, not it lowers the average cost of a surgical procedure. In
addition, in a pandemic such as the one caused by Covid-19, severe shortages of assistant surgeons and additional safety measures gravely limit the number of surgeries to be carried out, at a cost of thousands of valuable lives.

\subsection{The SARAS system}
The Smart Autonomous Robotic Assistant Surgeon
(SARAS) research project aims to replace the assistant surgeon with a pair of fully autonomous robotic arms. SARAS is designed to gain an edge over existing robotic-assisted systems by leveraging artificial intelligence (AI) to develop
a cognitive core capable of making the right decisions at the right time based on past observed events and its predictions of future surgical situations. SARAS is a complex system which involves technologies such as human-robot interaction, cognitive control, perception, navigation, and path planning. The aim of SARAS is to perform MIS procedures side by side with da Vinci, by interacting with the main surgeon in full autonomy. The system has therefore the potential to improve quality of life and patient safety, as well as economic efficiency for hospitals and healthcare systems.

\begin{figure}[h!]
\centering
\includegraphics[width=12cm, height=8cm]{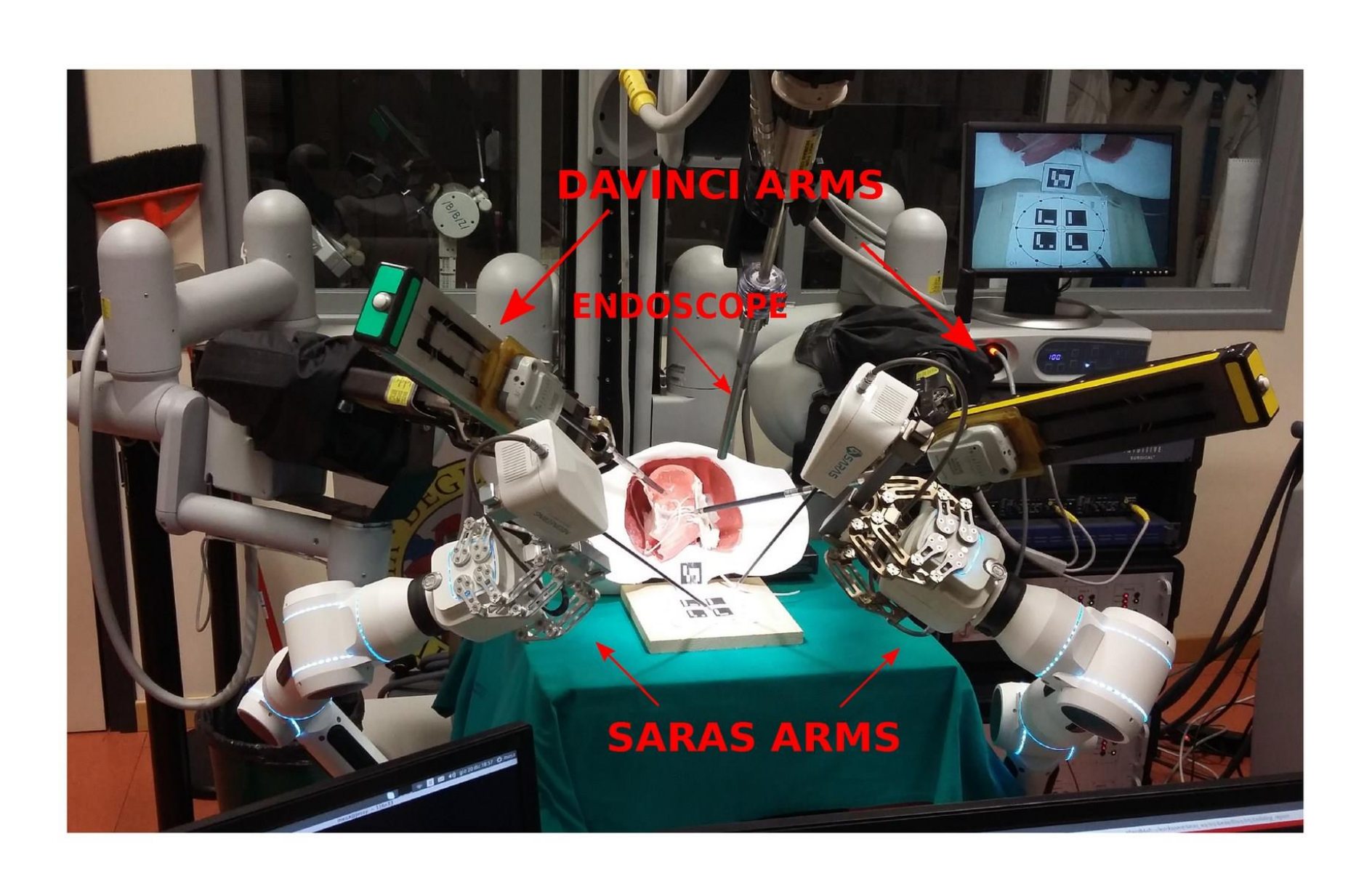}
\caption{The SARAS arms and a tele-operated da Vinci performing prostatectomy on an artificial, 3D printed anatomy (phantom).}
\label{fig1}
\end{figure}

The processing of visual input data captured by an endoscopic camera placed inside the surgical cavity in real-time plays a crucial role in informing the decision making of both the primary surgeon and the SARAS robotic arms, which need to work out the most appropriate action to put in place. Here, the human-robot interaction  in operating room is also vital for robotic-system to foresee the surgeon's intention to perceive suitable action \cite{PMID:33087597}. In this work we assume a configuration in which da Vinci acts as the main surgeon, while two SARAS robotic
arms play the role of the assistant surgeon, as shown in Figure \ref{fig1}. The perception component of SARAS involves various computer vision tasks, including online surgeon action recognition, current procedure stage recognition, predicting future surgeon actions, decision-making, and anomaly detection.

\subsection{Anomaly detection in endoscopic surgery}

\begin{figure}[h!]
\centering
\includegraphics[width=12cm, height=8cm]{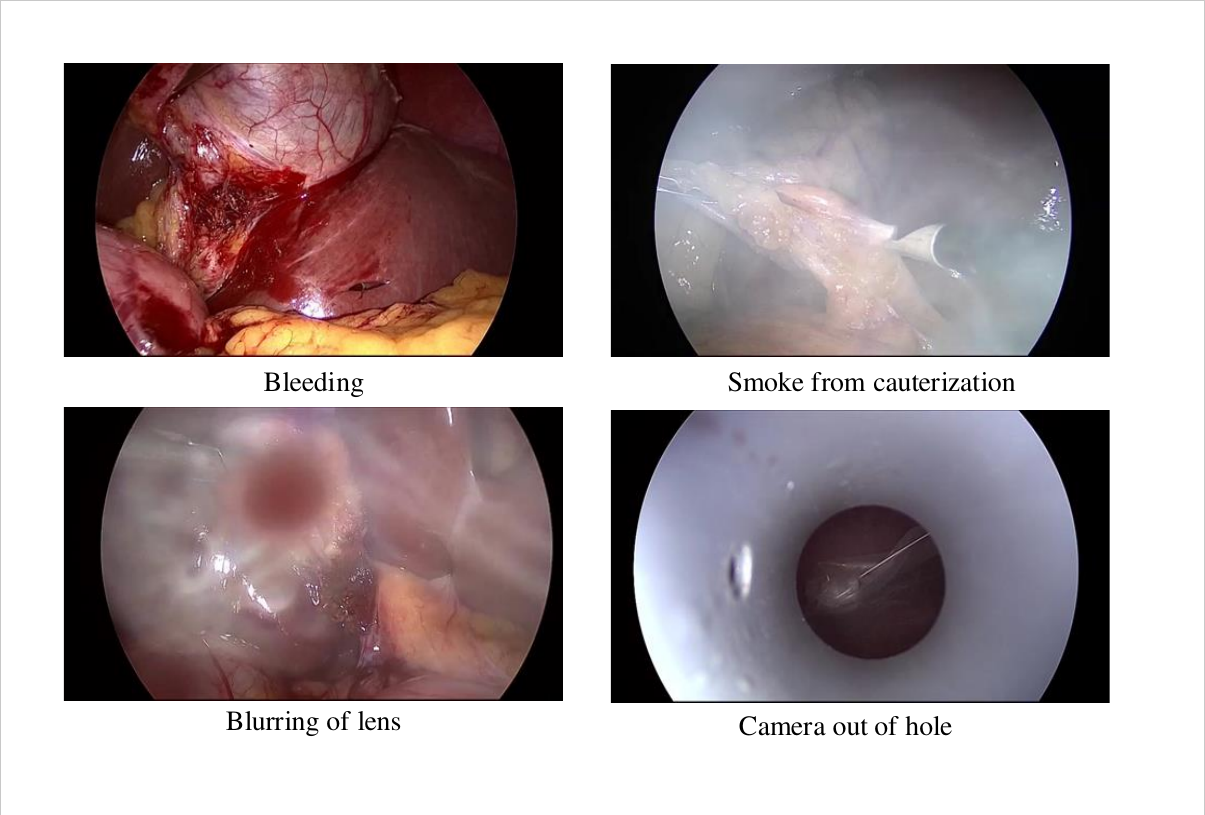}
\centering
\caption{Types of anomalies possibly occurring during cholecystectomy.}
\label{fig2}
\end{figure}

During endoscopic minimally-invasive surgery, a high-definition endoscopic camera is inserted into a hole in the patient’s body (trocar), and the surgeons observe the resulting video on a computer monitor. Based on the visuals from the endoscopic camera, the primary surgeon performs the surgery. Hence, a clear field of view of the surgical area is of paramount importance. Unfortunately, the field of view often
happens to be occluded in unexpected fashions. On account of the surgeon’s dependency on the camera feed, obstructions may moderately restrict the endoscopic feed’s usability. In general, occlusions or any event that deviates from the normal workflow of the procedure are called ’anomalies.’ Different types of anomalies exist, including bleeding, the presence of smoke due to surgical electro-cauterisation, the blurring of camera lenses, or the camera being out of the
trocar at some point during the surgery. Figure \ref{fig2} shows several example situations representing different anomalous scenes occurring during endoscopic surgery.

Existing methods for detecting anomalies in endoscopic data combine traditional image-based feature extraction with supervised learning \cite{6497444, 4760224,4548793,Xing2018BleedingDI}. Such models are trained using both available anomalous and non-anomalous data, and the hand-crafted features extracted from each video frame are used for classifying anomalies. Liu and Yuan, for instance, have employed colour-based feature extraction and a support vector machine (SVM) classifier to detect bleeding from endoscopic images acquired through Wireless Capsule Endoscopy (WCE) \cite{Liu2009ObscureBD}. Ghosh et al. have proposed a YIQ (Y-Luminance, IQ-Chrominance) color scheme for feature extraction from WCE videos. To differentiate bleeding from non-bleeding frames, statistical measures of the pixel values such mean, skewness, median, and minima are computed.
Finally, a standard SVM is employed for classification. In 2014, Sainju et al. have proposed statistical feature extraction for WCE images using first-order histogram probability of three RGB color spaces \cite{10.1007/s10916-014-0025-1}. In 2017, Okamoto et al. have designed a system for the real-time identification of bleeding based on a combination of RGB and HSV color values for feature extraction. The selected features are then classified in real time using, again, an SVM \cite{pub.1107333892}. Research on smoke detection has been rather limited in endoscopy, whereas it has attracted more interest in non-medical applications such as forest fire and surveillance smoke detectors. Leibetseder et al., in particular, have proposed a saturation analysis for
extracting features from the endoscopic frames and used an 8-layer AlexNet for classification \cite{10.1145/3126686.3126690}.

All such methods, however, go nowhere near as far as to meet the needs of real-time anomaly detection in R-MIS. The fundamental reason is that anomalous events of
any kind, rather than the most anticipated ones such as bleeding and smoke, can happen in autonomous robotic surgery, making any approach based on supervised learning unsuitable. Furthermore, to the best of our knowledge there is currently no accepted benchmark in anomaly detection in endoscopy, either supervised or unsupervised, as authors use different training/testing splits (e.g. a random selection of bleeding vs non-bleeding frames) and no dataset specifically
designed for this task exists.
\subsection{Rationale for unsupervised deep anomaly detection}

Real-time video anomaly detection in R-MIS is challenging due to the varied nature of different anomalies, the sparse occurrence of anomalous events, and the imbalance
between the amount of data available under normal and abnormal scenarios. As we have seen, existing approaches to anomaly detection in endoscopic videos are largely based on statistical analysis and/or supervised learning, which relies on the availability of labeled data for both anomalous and non-anomalous situations. Unfortunately, labelled data for anomalous events is unavailable and even hard to define in robotic-assisted surgery. A possible solution is provided by unsupervised learning, in particular in a deep learning formulation, which does not assume any annotation.

In particular, a class of deep neural networks known as deep autoencoders (DAE) has recently been proposed that is suitable for this task. These are reconstruction-based models which learn from ’normal’ data and detect anomalous frames which deviate from the normal visual pattern. An autoencoder consists of an encoder and a decoder. The encoder takes the input images/frames and reduces their
dimensionality by mapping them to a latent space. The resulting ’bottleneck’ features from this latent space are used by the decoder to minimise the error between original and reconstructed frames. Video frames associated with high reconstruction costs may thus be subject to anomaly \cite{Oluwatoyin2012VideoBasedAH}.

Some efforts on unsupervised video anomaly detection have been made in the wider computer vision field. E.g., Cho et al. have proposed an implicit autoencoder with a
SlowFast network structure for anomaly detection in surveillance videos \cite{cho2021unsupervised}. Recently, a robust unsupervised anomaly detection approach has been proposed by Wang et al. which employs a ConvGRU-based prediction network for capturing the spatiotemporal dependencies characterising normal data to predict anomalous frames \cite{wang2020robust}.

\subsection{Contributions}

In this paper we propose a deep autoencoder-based approach to real-time video anomaly detection in SARAS. The model learns spatial information from the input data
and strives to reconstruct the input video frames with low error. The proposed anomaly detection model for R-MIS is validated using two datasets: the publicly available Cholec80 dataset, adapted for anomaly detection, and a SARAS dataset
of videos capturing radical prostatectomy (RARP) procedures conducted on artificial anatomies (\textit{phantoms}), achieving extremely promising results. In summary

\begin{itemize}
    \item We propose the first deep unsupervised system for video anomaly detection in endoscopic surgery.
\item We contribute the first benchmarks for unsupervised surgical anomaly video detection, based on the existing Cholec80 dataset and our own SARAS dataset.
\item Our results show that our approach is able to detect all typical surgical video anomalies with high accuracy.
\item The system has been implemented and deployed in SARAS for video anomaly detection in real-time.
\end{itemize}
We plan to release our annotation upon acceptance to share
the new benchmarks with the community.

\section{DEEP RESIDUAL AUTOENCODERS}

In many real-world problems, obtaining significant samples of rare or anomalous situations is challenging. As a result, anomaly detection is forced to determine cases of abnormal instances by relying on training the system with only ‘normal’ or ’normal’ samples. Unsupervised anomaly detection then determines anomalous instances as those deviating from the distribution learned for the normal samples. Unsupervised learning approaches have recently gained momentum in computer vision, thanks for their not relying on expensive and time-consuming labeled datasets. Here
we propose in particular to tackle anomaly detection in an unsupervised approach based on deep residual autoencoders.

A deep residual autoencoder is a form of generative deep neural network, inspired by the discovery in neuroscience of shortcut connections in the brains of various animals, in turn emulated by residual convolutional neural networks \cite{article13, article14}.
The term residual learning relates to variables that consist of residual vectors between two segments of a long sequence. Residual vectors have been shown to be effective in learning shallow feature representations for image recognition tasks \cite{PMID:22156101, 10.1145/1179352.1142005}. Based on these facts, residual convolutional neural networks have been proposed for improving the accuracy of deep learning networks \cite{7780459, DBLP:journals/corr/HeZR016}.

The architecture consists of a continuous, stacked sequence of residual units, connected in a sequence with shortcut connections in a residual CNN. Each residual unit consists of three convolution layers (see Figure \ref{fig3}). The three sequentially-connected convolution layers collectively map an input $x$ to an output $F(x)$. The latter is then added to the input $x$ of the residual unit via a shortcut connection. Overall, the output of the residual unit can be expressed
as $H(x) = F(x) + x$. The condition $F(x) = 0$ indicates the disappearance of the network gradient weights, in which case $H(x) = x$ tends to an identity mapping that decreases the network’s depth while guaranteeing classification accuracy \cite{10.5555/3157096.3157158}. Biological findings in the brain show the pivotal role of similar hidden shortcut connections for synchronised motor movement, recovery from injuries, and reward learning. This is why the mechanism of shortcut connections has been injected into deep learning, especially in networks such as residual CNN and U-Net \cite{DBLP:journals/corr/RonnebergerFB15, 7780459}.

In this work we thus introduced residual connections in a traditional autoencoder architecture to implement deep residual learning for anomaly detection. Residual blocks are stacked in a balanced way in both the encoder and the decoder. The shallow layers of the encoder are connected to the decoder’s deep layers using shortcuts to encourage the formation of identity mappings. The forward and backward
error signals can be propagated between encoder and decoder using these identity mappings. Further, the use of rectified linear unit (ReLU) activation functions ensures faster convergence and propagation of error signals.

\section{METHODOLOGY}

The proposed methodology for anomaly detection is thus based on an unsupervised learning approach using a deep residual autoencoder. The latter learns from a dataset of ‘normal’ videos and then uses the learned parameters to identify abnormal behavior by thresholding the reconstruction error. The idea is that anomalous frames will have high reconstruction error compared to that of normal frames.

As mentioned above, a deep residual autoencoder architecture has residual connections implemented between layers of the encoder and the decoder. When compared
with the concatenated shortcuts in U-Net \cite{DBLP:journals/corr/RonnebergerFB15}, these residual connections have the property of minimising the number of training parameters of the model and of enhancing learning by propagating errors between the layers more efficiently. In this way the model is able to perform multi-level residual learning by combining dense residual blocks on different levels. Eventually, at the end of each residual block the feature maps outputted by each convolutional layer are concatenated.

\subsection{Architecture}

\begin{figure}[h]
\centering
\includegraphics[width = 8.5cm]{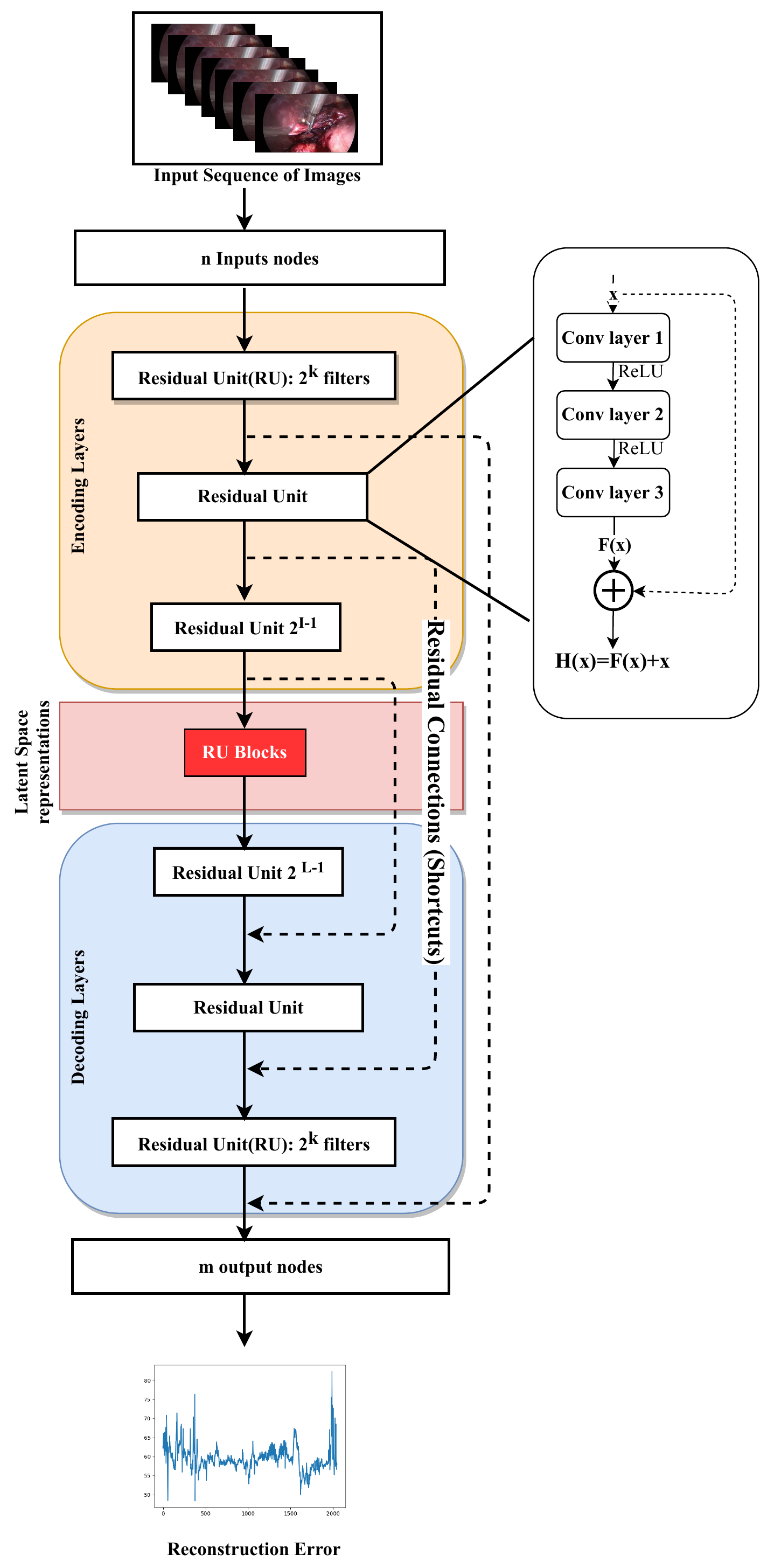}
\centering
\caption{Architecture of a deep residual autoencoder.}
\label{fig3} \vspace{-6mm}
\end{figure}

The typical deep residual autoencoder has $n$ inputs and $m$ target outputs with symmetrical hidden layers, as shown in Figure \ref{fig3}. The architecture also uses an element-wise rectified linear unit (ReLU) activation function and batch normalization (BN) for re-scaling in the network’s hidden layers to improve training. As explained, the residual connections in the encoder-decoder architecture address the exploding or vanishing gradient problems \cite{DBLP:journals/corr/SrivastavaGS15, DBLP:journals/corr/IoffeS15} and improve accuracy.
These residual connections help construct a mirrored identity mapping between the encoder and decoder layers in the network. Thus, in this work a nested residual convolutional network connected from the outermost to the innermost layers is implemented for anomaly detection.

Let $l$ denote an encoding layer and $L$ the corresponding decoding layer of the network. The input and output of the encoding layer $l$ are denoted by $X_l$ and $Y_l$, respectively, and by $X_L$ and $Y_L$ for the corresponding decoding layer $L$. The residual connection between corresponding layers mitigates the loss of information when back propagating losses during training \cite{rs11182142}. The relationship among the relevant quantities is illustrated in Equation \ref{eq1}:
\begin{equation} %FAB need to check the notation of this paragraph
\begin{split}
Y_L & = X_l + f_L(X_L,W_L)
%\\ & 
= X_l + f_L(g_L(f_l(X_l,W_l)), W_L),        
\end{split}
\label{eq1}
\end{equation}

where the activation functions for the encoding layer $l$ and the corresponding decoding layer $L$ are denoted by $f_l(X_l,W_l)$ and $f_L(X_L,W_L)$, respectively. Consequently, $X_L= g_L(f_l(X_l,W_l))$ represents the recursive mapping between the shallow encoder layer input $X_l$ and the deep decoder layer input $X_L$. The layers has $K$ size filters for convolutional operations. Deep autoencoders can efficiently learn an encoding from input data through dimensionality reduction via bottleneck features \cite{10.1016/j.neucom.2008.04.030, 10.1016/j.neucom.2013.09.055}. Our architecture for anomaly detection from videos in R-MIS has a symmetrical structure with the same number of hidden nodes and the same node dimensionality in both the encoder and decoder layers.

\subsection{Reconstruction error-based detection}

As explained above, in our approach anomaly detection is reconstruction error-based. The hypothesis is that, after the learning process, the autoencoder can reconstruct input frames never seen before, under the assumption that the latter resemble ’regular’ frames observed during training. Conversely, an autoencoder would struggle to reconstruct anomalous frames not matching the learnt feature maps. Therefore, abnormal frames would have a high reconstruction cost when compared to normal frames.

Our error measure is based on the relative difference between normal and abnormal structures, rather than its absolute value. %and not on the absolute cost value.   
To evaluate the performance of the system, we use a robust regressive loss function known as \emph{root mean squared error} (RMSE). The RMSE loss function heavily penalises reconstructed values which stray far from 'normal' values,
and is defined as:
\begin{equation}
        RMSE=\sqrt{\frac{\sum_{i=1}^{M}\sum_{j=1}^{N}\left(R\left[i,j\right]-F\left[i,j\right]\right)^2}{M\times N}}
\end{equation}
where $R[i,j]$ and $F[i,j]$ are the pixel values of the reference image coordinates and the reconstructed image, respectively.

\section{EXPERIMENTAL RESULTS}

Our residual autoencoder architecture was trained on a Quadro RTX 6000 8-GPU server with 24 GB VRAM per card. Our SARAS video anomaly detection system performance was evaluated using both anomalous and non-anomalous (normal) frames generated by the SARAS demonstration platform and an existing dataset in the surgery domain, Cholec80, adapted for anomaly detection. In our tests, each frame in the input video was passed to the autoencoder in order to measure the reconstruction error.

\subsection{Datasets}

The Cholec80 dataset was selected to create a suitable benchmark. It contains 80 videos of cholecystectomy procedures, performed by 13 different surgeons \cite{DBLP:journals/corr/TwinandaSMMMP16}. The videos are captured at 25 frames per second (fps) with a resolution of 854$\times$480 pixels, and contain both anomalous and non-anomalous frames. 
As it was designed to validate procedure phase recognition and tool detection approaches, rather than anomaly detection, in the Cholec80 dataset each frame is labelled with the phase of the procedure (as defined by a senior surgeon, at 25 fps) and the presence of tools (1 fps).

The model was also tested upon the SARAS platform prostatectomy dataset acquired through the SARAS SOLO-SURGERY system \cite{9078032}. The surgery was there performed on a prostate 3D-printed phantom by both the da Vinci and the SARAS robotic arms. The SARAS training dataset is composed by 18 video clips captured at 25 fps and 720$\times$480 resolution, containing only non-anomalous data (as anomalies are difficult to simulate in a phantom environment). SARAS videos also come with extra annotation, in the form of 23 relevant classes of surgeon actions, as in the real-world SARAS-ESAD surgical dataset used for a recent MIDL 2020 challenge on surgeon action detection \cite{bawa2020esad, bawa2021saras}. 

Note that although the SARAS dataset does not contains any \emph{surgical} anomalies, it is affected by a number of \emph{technical} anomalies, such as blocked views or loss of focus while recording the videos, which can actually mimic  anomalies one would realistically expect once the complete, fully automated robotic-assisted system is deployed.

\subsection{Training and Testing Protocol}

For our experiments we selected in each dataset a subset of frames for training and a separate, disjoint set of frames for testing. For Cholec80 we identified 65 ’normal’ video clips to train the model with normal data only. For testing, 3
video clips containing anomalies were used. In total, 5,584 frames were selected for testing and manually labelled. The model for the SARAS phantom dataset was trained upon a suitable selection of frames from the 18 video clips and tested on a total of 4,799 frames.

For training, the input frames were extracted and resized to $128\times128$ to be passed to the deep residual encoder-decoder pipeline. The deep residual autoencoder model was trained with a learning rate of 0.001 and a batch size of 64. The epoch delivering the lowest reconstruction error on average over the training samples was selected. In our experiments, the model converged at the 40th epoch on Cholec80 and 30th epoch on the SARAS dataset.

\subsection{ Anomaly Detection}

\begin{figure}[h]
\centering
\includegraphics[width=11cm, height=10cm]{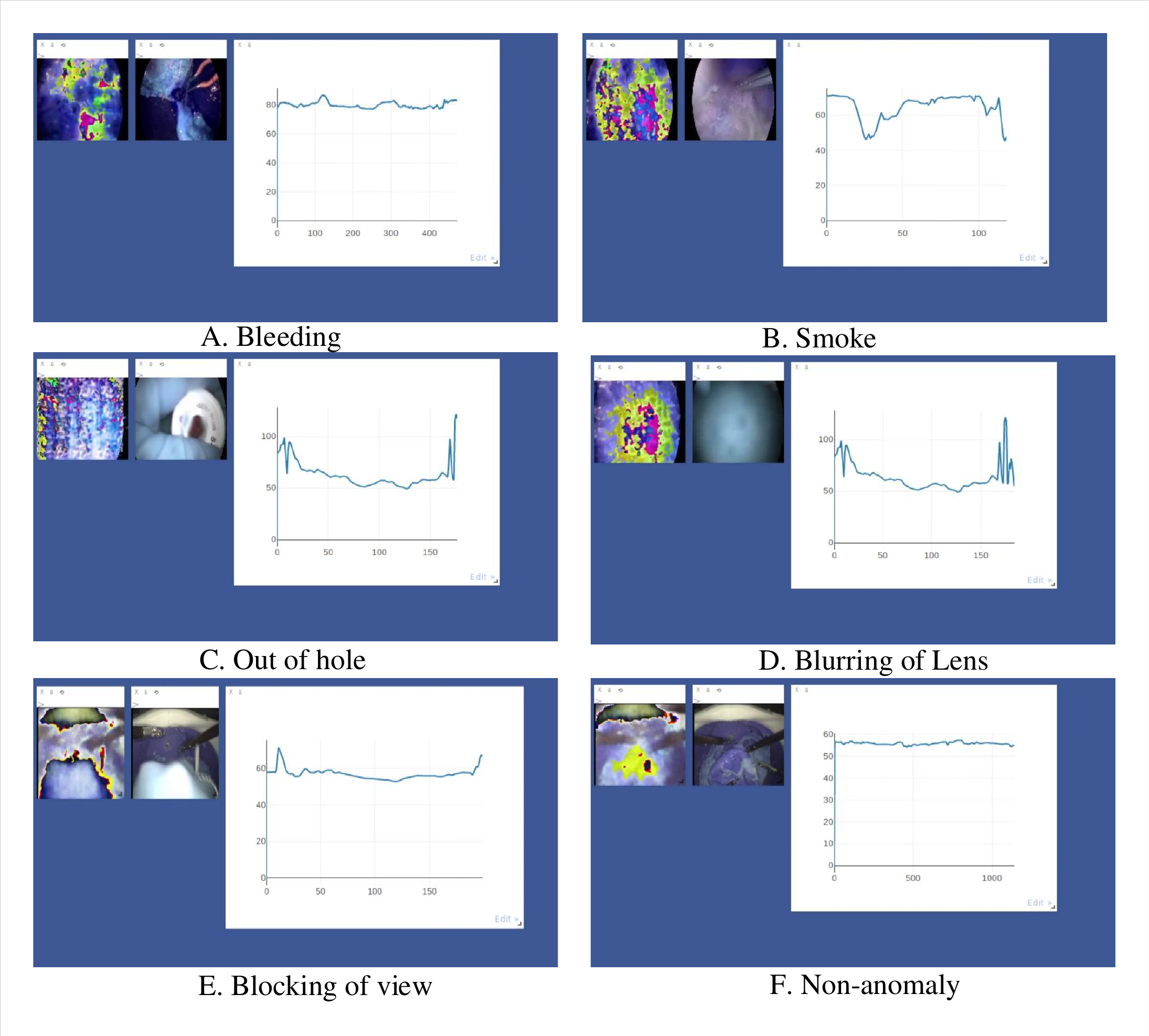}
\centering
\caption{A, B, C and D plot the reconstruction error over time and a pair of reconstructed/actual anomalous frames for four videos in the Cholec80 dataset. The X-axis represents the frames and Y-axis represents the reconstruction error. Different causes for the anomaly are considered. E and F do the same for two videos of the SARAS dataset, one containing an anomaly (blocked view, corresponding to spikes in the reconstruction error) and one not containing any.}
\label{fig4}
\end{figure}

Figure \ref{fig4} shows the interactive visualisation dashboard we deployed for live data experimentation, with both the reconstruction error graph generated by the trained autoencoder and the reconstructed images for some sample anomalous frames. The $X$-axis represents the frame number, while the $Y$-axis represents the reconstruction error value. To quantitatively discriminate anomalies and prompt SARAS to take precautionary actions, such as suction in case of bleeding or smoke or calling for manual intervention for other types of anomalies, we applied (on Cholec80) a lower threshold $\theta_{i\ }$  and an upper threshold $\theta_{j\ }$  to the reconstruction error. If the error for the current frame m was below $\theta_{i\ }$  or above $\theta_{j\ }$, we would flagged a possible anomaly, otherwise the frame was considered to be ’normal’. The reason for using a lower and an upper threshold is due to the effect of smoke, which tends to lower the value of the reconstruction error due to saturation of pixel intensities in the frame. As for the choice of the threshold, we adopted the $n-th$ percentile of the error distribution for normal frames.

After having an expert manually label a significant fraction of the video frames in the two datasets as either normal or anomalous, as already explained, our anomaly detection approach was evaluated by means of the usual precision and recall measures:
\begin{equation}
Recall = \frac{TP}{TP+FN}, \quad Precision = \frac{TP}{TP+FP}
\end{equation}
where TP is the number of true positives (anomalous frames), FP the number of false positives and FN that of false negatives (normal frames). Although anomaly detection is completely unsupervised, this labelling was conducted to allow a quantitative assessment of the system’s performance.
\begin{table}[h!]
\centering
\caption{Recall, Precision, F1-score over the test folds}
\begin{tabular}{|l|l|l|l|l|}
\hline
\textbf{Anomalies}          & \textbf{Total Frames} & \textbf{Recall} & \textbf{Precision} & \textbf{F1-Score} \\ \hline
\textbf{Cholec80 Overall}   & 5584                  & 78.4            & 91.5               & 84.4              \\ \hline
\textit{Bleeding}           & 2925                  & 76.6            & 93.7               & 84.2              \\ \hline
\textit{Smoke}              & 411                   & 100             & 89.2               & 94.2              \\ \hline
\textit{Camera out of hole} & 754                   & 76.6            & 89.2               & 71.3              \\ \hline
\textbf{SARAS dataset}      & 4799                  & 95.6            & 88.1               & 91.6              \\ \hline
\end{tabular}
\label{tab1}
\end{table}

\begin{table}[h!]
\centering
\caption{Overall confusion matrix for Cholec80 dataset}
\begin{tabular}{|l|p{5cm}|p{5cm}|}
\hline
\textbf{Prediction}           & \textbf{Frames detected as anomaly} & \textbf{Frames detected as non-anomaly} \\ \hline
\textbf{Anomalous frames}     & 3,743                         & 347                              \\ \hline
\textbf{Non-anomalous frames} & 1,031                         & 463                              \\ \hline
\end{tabular}
\label{tab2}
\end{table}

\subsection{Detection Accuracy}

The values of recall, precision and F1-score obtained are reported in Table \ref{tab1} for both Cholec80 and SARAS. The confusion matrix for the overall test data in the Cholec80 dataset is shown in Table \ref{tab2}. On Cholec80, our model detected anomalies with a recall of 78.4\% and a precision of 91.5\% using a threshold range of [$\theta_{i\ }$ = 48,$\theta_{j\ }$ = 65]. On SARAS, our model achieved a recall of 95.63\% and a precision of 88.10\%. Here anomalies were detected using a single threshold $\theta_{i\ }$ >= 57.4. Speed-wise, we achieved real-time testing performance on a single NVIDIA GTX 1070 GPU with 8 GB VRAM of 25ms per frame, for clips with frame resolution $854\times480$.
\section{CONCLUSIONS}

Overall, the proposed architecture has been shown capable of accurately detect anomalies in real-time in robotic-assisted systems in an unsupervised fashion. The model uses deep autoencoders with residual connections to propagate gradient
values throughout the network more efficiently. The complete system was validated on the real-data Cholec80 surgical dataset and the SARAS phantom dataset, achieving promising results for unsupervised anomaly detection in endoscopic videos in terms of both accuracy and speed. The lack of any previous unsupervised anomaly detection baselines on either Cholec80 or SARAS does not unfortunately allow us
to compare those results with prior work.

In the near future, we plan to extend this approach to detect anomalies using 3D video feature extraction in a Generative Adversarial Network (GAN) architecture.

\bibliographystyle{unsrt}  
\bibliography{references}  %%% Remove comment to use the external .bib file (using bibtex).

\begin{thebibliography}{10}

\bibitem{Guthart2000TheIT}
G.~Guthart and J.~Salisbury.
\newblock The intuitive/sup tm/ telesurgery system: overview and application.
\newblock {\em Proceedings 2000 ICRA. Millennium Conference. IEEE International
  Conference on Robotics and Automation. Symposia Proceedings (Cat.
  No.00CH37065)}, 1:618--621 vol.1, 2000.

\bibitem{Vitiello2013EmergingRP}
V.~Vitiello, Su-Lin Lee, T.~Cundy, and G.~Yang.
\newblock Emerging robotic platforms for minimally invasive surgery.
\newblock {\em IEEE Reviews in Biomedical Engineering}, 6:111--126, 2013.

\bibitem{LEE201033}
Su-Lin Lee, Mirna Lerotic, Valentina Vitiello, Stamatia Giannarou, Ka-Wai Kwok,
  Marco Visentini-Scarzanella, and Guang-Zhong Yang.
\newblock From medical images to minimally invasive intervention: Computer
  assistance for robotic surgery.
\newblock {\em Computerized Medical Imaging and Graphics}, 34(1):33--45, 2010.
\newblock Image-Guided Surgical Planning and Therapy.

\bibitem{PMID:33087597}
Inna Skarga-Bandurova, Rostislav Siriak, Tetiana Biloborodova, Fabio Cuzzolin,
  Vivek~Singh Bawa, Mohamed~Ibrahim Mohamed, and R~Dinesh~Jackson Samuel.
\newblock Surgical hand gesture prediction for the operating room.
\newblock {\em Studies in health technology and informatics}, 273:97—103,
  September 2020.

\bibitem{6497444}
Y.~{Fu}, W.~{Zhang}, M.~{Mandal}, and M.~Q.~. {Meng}.
\newblock Computer-aided bleeding detection in wce video.
\newblock {\em IEEE Journal of Biomedical and Health Informatics},
  18(2):636--642, 2014.

\bibitem{4760224}
B.~{Li} and M.~Q.~. {Meng}.
\newblock Computer-aided detection of bleeding regions for capsule endoscopy
  images.
\newblock {\em IEEE Transactions on Biomedical Engineering}, 56(4):1032--1039,
  2009.

\bibitem{4548793}
Y.~S. {Jung}, Y.~H. {Kim}, D.~H. {Lee}, and J.~H. {Kim}.
\newblock Active blood detection in a high resolution capsule endoscopy using
  color spectrum transformation.
\newblock In {\em 2008 International Conference on BioMedical Engineering and
  Informatics}, volume~1, pages 859--862, 2008.

\bibitem{Xing2018BleedingDI}
Xiaohan Xing, Xiao Jia, and M.~Meng.
\newblock Bleeding detection in wireless capsule endoscopy image video using
  superpixel-color histogram and a subspace knn classifier.
\newblock {\em 2018 40th Annual International Conference of the IEEE
  Engineering in Medicine and Biology Society (EMBC)}, pages 1--4, 2018.

\bibitem{Liu2009ObscureBD}
J.~Liu and X.~Yuan.
\newblock Obscure bleeding detection in endoscopy images using support vector
  machines.
\newblock {\em Optimization and Engineering}, 10:289--299, 2009.

\bibitem{10.1007/s10916-014-0025-1}
Sonu Sainju, Francis~M. Bui, and Khan~A. Wahid.
\newblock Automated bleeding detection in capsule endoscopy videos using
  statistical features and region growing.
\newblock 38(4), 2014.

\bibitem{pub.1107333892}
Takayuki Okamoto, Takashi Ohnishi, Hiroshi Kawahira, Olga Dergachyava, Pierre
  Jannin, and Hideaki Haneishi.
\newblock Real-time identification of blood regions for hemostasis support in
  laparoscopic surgery.
\newblock {\em Signal, Image and Video Processing}, 13(2):405--412, 2019.

\bibitem{10.1145/3126686.3126690}
Andreas Leibetseder, Manfred~J\"{u}rgen Primus, Stefan Petscharnig, and Klaus
  Schoeffmann.
\newblock Real-time image-based smoke detection in endoscopic videos.
\newblock Thematic Workshops '17, page 296–304, New York, NY, USA, 2017.
  Association for Computing Machinery.

\bibitem{Oluwatoyin2012VideoBasedAH}
P.~Oluwatoyin and K.~Wang.
\newblock Video-based abnormal human behavior recognition—a review.
\newblock {\em IEEE Transactions on Systems, Man, and Cybernetics, Part C
  (Applications and Reviews)}, 42:865--878, 2012.

\bibitem{cho2021unsupervised}
MyeongAh Cho, Taeoh Kim, Ig-Jae Kim, and Sangyoun Lee.
\newblock Unsupervised video anomaly detection via normalizing flows with
  implicit latent features, 2021.

\bibitem{wang2020robust}
Xuanzhao Wang, Zhengping Che, Ke~Yang, Bo~Jiang, Jian Tang, Jieping Ye, Jingyu
  Wang, and Qi~Qi.
\newblock Robust unsupervised video anomaly detection by multi-path frame
  prediction, 2020.

\bibitem{article13}
Arpiar Saunders, Ian Oldenburg, Vladimir Berezovskii, Caroline Johnson, Nathan
  Kingery, Hunter Elliott, Tiao Xie, Charles Gerfen, and Bernardo Sabatini.
\newblock A direct gabaergic output from the basal ganglia to frontal cortex.
\newblock {\em Nature}, 521:85--89, 03 2015.

\bibitem{article14}
Moriel Zelikowsky, Stephanie Bissiere, Timothy Hast, Rebecca Bennett, Andrea
  Cowley, Bryce Vissel, and Michael Fanselow.
\newblock Prefrontal microcircuit underlies contextual learning after
  hippocampal loss.
\newblock {\em Proceedings of the National Academy of Sciences of the United
  States of America}, 110, 05 2013.

\bibitem{PMID:22156101}
Hervé Jégou, Florent Perronnin, Matthijs Douze, Jorge Sánchez, Patrick
  Pérez, and Cordelia Schmid.
\newblock Aggregating local image descriptors into compact codes.
\newblock {\em IEEE transactions on pattern analysis and machine intelligence},
  34(9):1704—1716, September 2012.

\bibitem{10.1145/1179352.1142005}
Richard Szeliski.
\newblock Locally adapted hierarchical basis preconditioning.
\newblock SIGGRAPH '06, page 1135–1143, New York, NY, USA, 2006. Association
  for Computing Machinery.

\bibitem{7780459}
K.~He, X.~Zhang, S.~Ren, and J.~Sun.
\newblock Deep residual learning for image recognition.
\newblock In {\em 2016 IEEE Conference on Computer Vision and Pattern
  Recognition (CVPR)}, pages 770--778, Los Alamitos, CA, USA, jun 2016. IEEE
  Computer Society.

\bibitem{DBLP:journals/corr/HeZR016}
Kaiming He, Xiangyu Zhang, Shaoqing Ren, and Jian Sun.
\newblock Identity mappings in deep residual networks.
\newblock {\em CoRR}, abs/1603.05027, 2016.

\bibitem{10.5555/3157096.3157158}
Andreas Veit, Michael Wilber, and Serge Belongie.
\newblock Residual networks behave like ensembles of relatively shallow
  networks.
\newblock NIPS'16, page 550–558, Red Hook, NY, USA, 2016. Curran Associates
  Inc.

\bibitem{DBLP:journals/corr/RonnebergerFB15}
Olaf Ronneberger, Philipp Fischer, and Thomas Brox.
\newblock U-net: Convolutional networks for biomedical image segmentation.
\newblock {\em CoRR}, abs/1505.04597, 2015.

\bibitem{DBLP:journals/corr/SrivastavaGS15}
Rupesh~Kumar Srivastava, Klaus Greff, and J{\"{u}}rgen Schmidhuber.
\newblock Highway networks.
\newblock {\em CoRR}, abs/1505.00387, 2015.

\bibitem{DBLP:journals/corr/IoffeS15}
Sergey Ioffe and Christian Szegedy.
\newblock Batch normalization: Accelerating deep network training by reducing
  internal covariate shift.
\newblock {\em CoRR}, abs/1502.03167, 2015.

\bibitem{rs11182142}
Lianfa Li.
\newblock Deep residual autoencoder with multiscaling for semantic segmentation
  of land-use images.
\newblock {\em Remote Sensing}, 11(18), 2019.

\bibitem{10.1016/j.neucom.2008.04.030}
Cheng-Yuan Liou, Jau-Chi Huang, and Wen-Chie Yang.
\newblock Modeling word perception using the elman network.
\newblock {\em Neurocomput.}, 71(16–18):3150–3157, October 2008.

\bibitem{10.1016/j.neucom.2013.09.055}
Cheng-Yuan Liou, Wei-Chen Cheng, Jiun-Wei Liou, and Daw-Ran Liou.
\newblock Autoencoder for words.
\newblock {\em Neurocomput.}, 139:84–96, September 2014.

\bibitem{DBLP:journals/corr/TwinandaSMMMP16}
Andru~Putra Twinanda, Sherif Shehata, Didier Mutter, Jacques Marescaux, Michel
  de~Mathelin, and Nicolas Padoy.
\newblock Endonet: {A} deep architecture for recognition tasks on laparoscopic
  videos.
\newblock {\em CoRR}, abs/1602.03012, 2016.

\bibitem{9078032}
A.~{Leporini}, E.~{Oleari}, C.~{Landolfo}, A.~{Sanna}, A.~{Larcher},
  G.~{Gandaglia}, N.~{Fossati}, F.~{Muttin}, U.~{Capitanio}, F.~{Montorsi},
  A.~{Salonia}, M.~{Minelli}, F.~{Ferraguti}, C.~{Secchi}, S.~{Farsoni},
  A.~{Sozzi}, M.~{Bonfè}, N.~{Sayols}, A.~{Hernansanz}, A.~{Casals},
  S.~{Hertle}, F.~{Cuzzolin}, A.~{Dennison}, A.~{Melzer}, G.~{Kronreif},
  S.~{Siracusano}, F.~{Falezza}, F.~{Setti}, and R.~{Muradore}.
\newblock Technical and functional validation of a teleoperated multirobots
  platform for minimally invasive surgery.
\newblock {\em IEEE Transactions on Medical Robotics and Bionics},
  2(2):148--156, 2020.

\bibitem{bawa2020esad}
Vivek~Singh Bawa, Gurkirt Singh, Francis KapingA, Inna Skarga-Bandurova, Alice
  Leporini, Carmela Landolfo, Armando Stabile, Francesco Setti, Riccardo
  Muradore, Elettra Oleari, and Fabio Cuzzolin.
\newblock Esad: Endoscopic surgeon action detection dataset, arxiv preprint
  arXiv:2006.07164, 12 Jun 2020.

\bibitem{bawa2021saras}
Vivek~Singh Bawa, Gurkirt Singh, Francis KapingA, Inna Skarga-Bandurova,
  Elettra Oleari, Alice Leporini, Carmela Landolfo, Pengfei Zhao, Xi~Xiang,
  Gongning Luo, Kuanquan Wang, Liangzhi Li, Bowen Wang, Shang Zhao, Li~Li,
  Armando Stabile, Francesco Setti, Riccardo Muradore, and Fabio Cuzzolin.
\newblock The saras endoscopic surgeon action detection (esad) dataset:
  Challenges and methods, arxiv preprint arXiv:2104.03178 2021.

\end{thebibliography}

%\begin{thebibliography}{1}

%\end{thebibliography}

\end{document}